\documentclass[9pt,twocolumn,twosides]{pnas-new}

\templatetype{pnasresearcharticle}

\usepackage{fancyhdr}
\usepackage{pdfpages}

\usepackage[T1]{fontenc}
\usepackage{ulem}
\usepackage{bm}
\usepackage{amsmath}
\usepackage{comment}

\DeclareMathOperator*{\argmax}{arg\,max}

\begin{document}

\title{Olfactory pursuit: catching a moving odor source in complex flows}

\author[a,b]{Maurizio Carbone}
\author[b]{Lorenzo Piro}
\author[c]{Robin A. Heinonen}
\author[b]{Luca Biferale}
\author[a,1]{Massimo Cencini}
\author[d,e,2]{Antonio Celani}

\affil[a]{Istituto dei Sistemi Complessi, CNR, Via dei Taurini 19, 00185 Rome, Italy and INFN ``Tor Vergata", Via della Ricerca Scientifica 1, 00133 Rome, Italy}
\affil[b]{Department of Physics \& INFN, Tor Vergata University of Rome, Via della Ricerca Scientifica 1, 00133 Rome, Italy}
\affil[c]{Machine Learning Genoa Center (MaLGa) \& Department of Civil, Chemical and Environmental Engineering, University of Genoa, Genoa, Italy}
\affil[d]{Quantitative Life Sciences, The Abdus Salam International Centre for Theoretical Physics, 34151 Trieste, Italy}
\affil[e]{Department of Oncology, University of Torino, Torino, Italy}
\leadauthor{Carbone}

\significancestatement{Tracking a moving target in a complex fluid environment while using sparse olfactory cues is a fundamental challenge shared by animal predators and autonomous sensing systems. While most prior work has focused on stationary odor sources, real-world targets often move strategically, leaving delayed and misleading chemical traces that are strongly mixed by a turbulent flow. We show that effective olfactory pursuit requires predicting target motion and inferring the turbulent dispersion properties rather than merely reducing uncertainty. When the target prey exhibits persistent motion, a planning-based, greedy strategy that anticipates future trajectories substantially outperforms purely exploratory approaches such as Infotaxis. Optimal behavior emerges from a hybrid strategy that dynamically balances information acquisition with exploitation of predicted motion, especially in regimes where predator and prey move at comparable speeds. These results clarify how cognition and prediction enhance search efficiency in turbulent-like, information-poor environments.}

\authorcontributions{M~Carbone: L~Piro: R~Heinonen: L~Biferale: M~Cencini: A~Celani: }
\authordeclaration{The authors declare no competing interests.}
\correspondingauthor{\textsuperscript{1}massimo.cencini@cnr.it, \textsuperscript{2}acelani@ictp.it}


\begin{abstract}
Locating and intercepting a moving target from possibly delayed, intermittent sensory signals is a paradigmatic problem in decision-making under uncertainty, and a fundamental challenge for, e.g., animals seeking prey or mates and autonomous robotic systems. Odor signals are intermittent, strongly mixed by turbulent-like transport, and typically lag behind the true target position, thereby complicating localization. Here, we formulate olfactory pursuit as a partially observable Markov decision process in which an agent maintains a joint belief over the target’s position and velocity. Using a discrete run-and-tumble model, we compute quasi-optimal policies by numerically solving the Bellman equation and benchmark them against well-established information-theoretic strategies such as Infotaxis. We show that purely exploratory policies are near-optimal when the target frequently reorients, but fail dramatically when the target exhibits persistent motion. We thus introduce a computationally efficient hybrid policy that combines the information-gain drive of Infotaxis with a ``greedy'' value function derived from an associated fully observable control problem. Our heuristic achieves near-optimal performance across all persistence times and substantially outperforms purely exploratory approaches. Moreover, our proposal demonstrates strong robustness even in more complex search scenarios, including {continuous} run-and-tumble prey motion with moderate persistence time, model mismatch, {and more accurate} plume dynamics {representation}. Our results identify predictive inference of target motion as the key ingredient for effective olfactory pursuit and provide a general framework for search in information-poor, dynamically evolving environments.
\end{abstract}

\dates{This manuscript was compiled on \today}

\maketitle
\thispagestyle{firststyle}
\ifthenelse{\boolean{shortarticle}}{\ifthenelse{\boolean{singlecolumn}}{\abscontentformatted}{\abscontent}}{}

\firstpage[1]{4}

\dropcap{S}earching for a mobile target using olfactory cues is a task relevant to fields ranging from behavioral ecology to autonomous robotics. In biological contexts, predators such as sharks or wolves must track prey that are not only distant but actively evading detection \cite{conover2007predator}, similarly swimming planktonic organisms need to find their mates or prey using their "scents" \cite{seuront2013chemical,yen2011pheromone}. In engineering contexts, autonomous underwater vehicles or aerial drones are increasingly tasked with locating leaking chemical sources in dynamic conditions \cite{kashyap2017pursuing,demetriou2013coupled}. Unlike searching for a stationary object, in which the target remains fixed while the searcher navigates the odor plume or trail, tracking a moving source introduces an additional layer of complexity. 

Previous studies mainly focused on the olfactory search for a stationary target \cite{balkovsky2002olfactory,Vergassola2007,kowadlo2008robot,chen2019odor}. When the searcher has prior knowledge about the likelihood of odor encounters, it can use a cognitive model of the target location, often in the form of a coarse-grained map\cite{reddy2022olfactory,masson2013olfactory,verano2023olfactory}. This approach frames the search as a partially observable Markov decision process (POMDP), enabling decision-making under uncertainty \cite{kaelbling1998}. By maintaining a belief map---a probabilistic estimate of the target's location---the agent can predict its position and act accordingly. 
In common situations where odor cues are scarce, such as in turbulent environments \cite{moore2004odor,celani2014odor}, the most effective strategies require an active search for information about the source location. As a matter of fact, strategies that only prioritize information acquisition, such as Infotaxis \cite{Vergassola2007}, have been shown to be high-performing in many settings in which the target source is stationary \cite{loisy2022,heinonen2023}.

\begin{figure*}
    \centering
    \includegraphics[width=1.\textwidth]{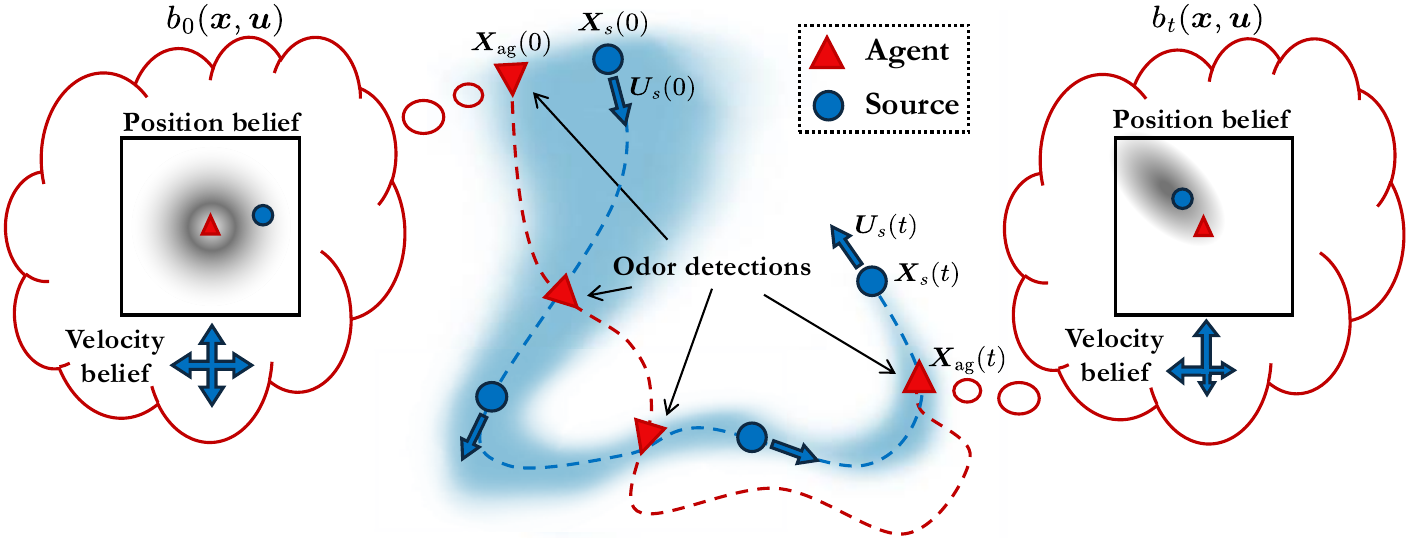}
    \caption{Sketch of typical olfactory pursuit. The target source (blue circle) performs a run-and-tumble motion and emits a blue chemical plume along its trajectory. The agent (red dashed trajectory) pursues the moving target by inferring its position and velocity from discrete odor detections (indicated by the red triangles). The agent maintains a probability map over the target position $\bm{X}_s$ and velocity $\bm{U}_s$, namely the joint belief $b_t(\bm{x},\bm{u})$ about its state at time $t$ represented in the thought clouds. The search begins with an initial detection at $t=0$, after which the agent’s position belief corresponds to the likelihood of a detection centered at the agent's position; the estimated target velocity is initially random uniform over all directions. As time progresses, the target moves and emits, while the agent updates its belief to account for the target source's motion and accumulates information about its  state through successive detections. Over time, the agent's belief becomes more peaked around the true target position, and its guess of the target velocity gets more accurate. This allows the agent to confidently pursue and intercept the target.}
    \label{fig_sketch}
\end{figure*}

When the target is in motion, we speak instead of \textit{olfactory pursuit}. In this case, the efficient searcher must also adopt a model of the target dynamics, and the belief map should account for both the position and velocity of the target (as sketched in Fig.~\ref{fig_sketch}). This endows the predator with the ability to anticipate the prey's movements and successfully capture it \cite{richardson2018unpredictable,szopa2022responsive}.
Such predictive capability is essential because it allows the agent to distinguish between the lingering ``ghost'' of an odor plume and the future trajectory of its target. Consequently, information-theoretic strategies must be adapted to account for the target's displacement.

In this paper, we demonstrate that when a target odor source moves with high persistence, maintaining a straight path for extended periods, a predator that plans ahead using a greedy strategy significantly outperforms one that merely explores. This suggests that the optimal strategy is often a hybrid one~\cite{Cassandra1996}: balancing the ``Infotaxis'' drive to reduce uncertainty with a ``Greedy'' drive to exploit the predicted path of the prey. This balance, which also proved to be useful for stationary targets~\cite{masson2013olfactory,loisy2022}, is particularly vital in the ``blind spot'' where the prey moves at a speed comparable to that of the predator.

To achieve this balance, we propose a composite policy in which the agent weighs the benefits of belief entropy reduction against the estimated value of immediate actions. While the ``Infotaxis'' component drives the agent to narrow down the search area by maximizing information gain from observations~\cite{Vergassola2007}, the ``Greedy'' component penalizes the expected time to capture by utilizing a value function solving the underlying Markov decision process (MDP) where the prey's instantaneous position and direction of motion are known, discouraging purely exploratory moves that would not lead to interception of the prey. This distinction is crucial because a purely explorative agent might systematically track the olfactory trail left by the target without ever committing to a capture trajectory, whereas a purely greedy agent risks simply following false leads generated by the stochasticity of detections. By adjusting the weight between these two drives, the searching agent can adapt its behavior to the target persistence time: relying on information-gathering when the prey frequently turns and switching to a direct interception course when the prey's ballistic trajectory becomes predictable.

The superiority of this predictive framework is demonstrated both in discrete lattice models and in a more complex continuous domain, in which the target dynamics are only approximately represented and the governing physical processes of odor dispersion give rise to an unsteady wake.

The supremacy of this predictive approach is validated both in discrete lattice models and in a more complex continuous environment, in which the agent has only approximate knowledge of the target dynamics and the olfactory cues are distributed in an unsteady wake. Indeed, in realistic scenarios, the odor plume steers and deforms with the target's motion, causing the chemical signal to lag behind the target's actual position, as illustrated in Fig.~\ref{fig_sketch}.
Our findings show that the hybrid heuristic remains robust in \sout{ realistic} environments where the prey’s directional-change timescale is comparable to the diffusion timescale of the chemical signal.

\section*{Results}

We address the olfactory pursuit problem through two strategic setups. We first consider a paradigmatic scenario in which both the agent and the target move on a lattice, with the target performing a discrete run-and-tumble. In this setting, we show two results. First, it is possible to fix a benchmark by numerically finding an approximate solution to the Bellman optimality equation \cite{bellman1957} for POMDPs \cite{astrom1965,sondik1978,kaelbling1998}. Second, we show that our proposed heuristic blending of pure exploitative and strongly explorative Bayesian policies achieves performances close to the Bellman solution, providing a robust and interpretable set of strategies to catch a moving target. Furthermore, to probe generalizability, we extend the problem to a more {complex} setup where the target moves off-lattice via continuous run-and-tumble dynamics. Also in this case, we show that the proposed heuristic policy continues to yield good performance, supporting its generalization to {highly representative} situations, where solving the Bellman equation is infeasible due to the curse of dimensionality and the lack of a precise model for either the target motion or the olfactory environment. 

\subsection*{Discrete run-and-tumble}
In this setup, agent and target move on a discrete lattice of spacing $\Delta x$ at discrete time steps $\Delta t$, both set to unity without loss of generality.
The target can move to its four nearest-neighbor lattice sites, and its direction of motion persists on average for a time $\tau_p$, after which it stops and selects a new direction (details on the model can be found in Methods). Along its trajectory, the target emits odor cues that are assumed to rapidly mix due to turbulent diffusion. These cues are drawn from the likelihood obtained from the exact solution of the diffusion equation for a stationary source. This assumption implies that the target is much slower than the characteristic odor diffusion speed (see Methods).

 At time $t$, the agent perceives a binarized olfactory signal $H(t)\in \{\texttt{detection},\texttt{no-detection}\}$ (odor concentration above/below a threshold of sensitivity). By assumption, the agent knows the likelihood of making a detection in its position, $\bm{X}_{\rm ag}(t)$, as a function $\mu$ of the distance from the target position $\bm{X}_s(t)$:
\begin{equation}
    \label{eq:likelihood}
    \mathcal{L}(\texttt{detection}| \bm{X}_{\rm ag} - \bm{X}_s) = 1 - \exp(-\mu(\bm{X}_{\rm ag}-\bm{X}_s))\,,
\end{equation}
where $\mu$ is the mean number of particles detected in a unit time (see Methods.
Furthermore, the agent has full knowledge of the probability, $P(\bm{u}'|\bm{u})$, of the target velocity $\bm{u}'=\bm{U}_s(t)$ at time $t$ conditioned on the previous velocity $\bm{u}=\bm{U}_s(t-1)$.
The information the agent has about the state of the moving target is encoded in the belief $b_t(\bm{x},\bm{u})$, namely a probability map about the state of the target, consisting of its actual position and velocity $[\bm{X}_s,\bm{U}_s]$. The belief $b_t$ encodes the agent's knowledge of the target's state, accounting for all detections up to time $t$.

The search always begins with an odor detection, so that the initial belief over position and velocity is the product of the likelihood \ref{eq:likelihood} and the invariant distribution of the target velocities, since at the beginning the agent does not possess any information on the target velocity (see Methods). 
The target is considered found when its distance from the agent  is smaller than the agent range of sight  (hereafter set to $\sqrt{2}$ in lattice spacing-units). The agent's goal is to catch the target, minimizing the (average) search time. This corresponds to finding a policy $\pi$ that maps the belief to an ``optimal'' action $a_t=\pi[b_t]$, where the allowed agent action consists of choosing one of the four adjacent grid points.

During its search, the agent updates its belief on the target position and velocity in two steps. First, using the transition probability $P$, the belief is transported to account for the target's motion. 
Second, the agent takes an action according to the prescribed policy and, if it does not find the moving target, records the presence of odor particles at its new position. Then, conditioned on the detection outcome $H(t)$, updates its belief about the target state using the likelihood~\ref{eq:likelihood} and Bayes’ rule, $b_{t+1} = \mathcal{B}[b_t,H(t)]$. Further details of this algorithmic procedure are provided in the Methods and Supplemental Material (SM).
The above-described setting is visualized in the supplementary movies S1-S2.

The discrete search on a lattice can be formulated as a solvable POMDP for which the target's state is defined via the agent's belief $b$ \cite{kaelbling1998}.
The POMDP is associated with a Bellman recurrence equation for the belief-state value function $V[b]$, i.e., the expected total reward \cite{krishnamurthy2016partially,spaan2012pomdp}:
\begin{subequations}
\begin{align}
\label{eq_bellman1}
V[b] &= \max_{\bm{a}} Q[b,\bm{a}] \\
\label{eq_bellman2}
Q[b, \bm{a}] &= \sum_{\bm{x},\bm{u}} b \mathcal{R} + \gamma \sum_{b'} \mathrm{Pr}(b'|b,\bm{a}) V[b'] \, ,
\end{align}
\label{eq_bellman}
\end{subequations}
where the second equation defines the state-action value function $Q[b(\bm{x},\bm{u}),\bm{a}]$ . In Eq.~\ref{eq_bellman1}, $\mathcal{R}(\bm{x},\bm{u},\bm{a})$ is the immediate reward gained, $\gamma\in(0,1)$ a discount factor, weighting between myopic and far-sighted planning, $\mathrm{Pr}[b'|b,\bm{a}]$ is the probability of transitioning from belief $b$ to $b'$ by taking action $\bm{a}$, comprising Bayes' rule to update $b'$ based on odor detections and target motion.
We set the reward to be unity upon finding the source, and zero otherwise (other choices are equivalent, provided $\gamma < 1$).
During the search, at each time $t$, the agent chooses the action $\bm{A}(t)$ that leads to the expected maximum value function given the belief $b_t$, as defined in Eq.~\ref{eq_bellman1}. As in Refs.~\cite{loisy2023,heinonen2025a,heinonen2025b}, we used the SARSOP algorithm~\cite{sarsop} to obtain an approximate representation of the belief-state value function, thereby yielding a quasi-optimal policy. Further details on the POMDP setup are provided in Methods.

\begin{figure*}[t!]
    \centering
    \includegraphics[width=1.\textwidth]{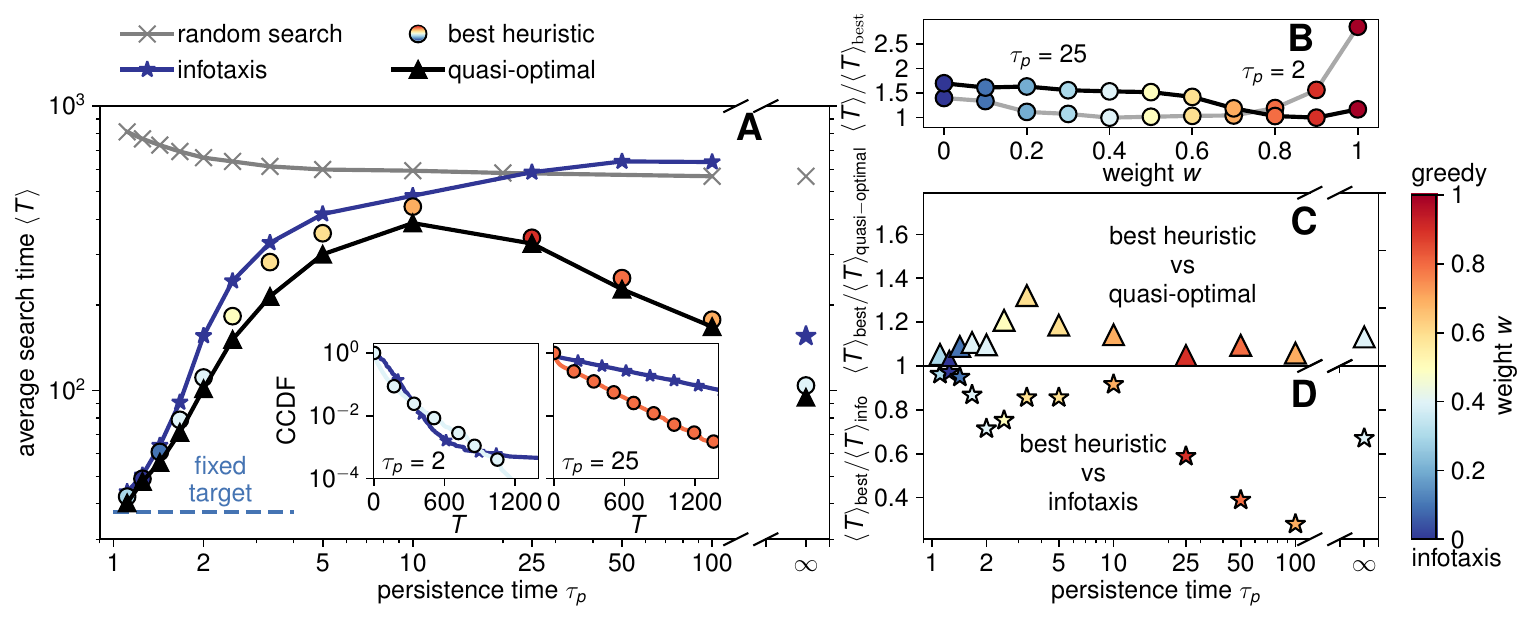}
    \caption{
    Results from the discrete run-and-tumble model, with both agent and target moving on a periodic lattice. Panel \textbf{A} shows the average search time $\langle T \rangle$ as a function of the target persistence time $\tau_p$, averaged over $10^4$ search episodes starting from a detection: the black curve indicates the quasi-optimal strategy obtained from solving the POMDP; blue curve the infotactic policy ($w=0$); gray curve refers to random search, obtained with the agent performing a persistent random walk with no olfactory cues; circles indicate the hybrid heuristic policy \ref{eq:blend} for the optimal weight $w$, yielding the lowest average search time, color-coded as in the right bar. The insets show the complementary cumulative distribution function (CCDF) of the search times for the best hybrid heuristic and Infotaxis, at $\tau_p=2$ and $25$.
    Panel \textbf{B} details the effect of the policy weight $w$ on the average search time, normalized with the best average search time, for $\tau_p=2$ and $25$. Panels \textbf{C},\textbf{D} compare the average search times of the best hybrid heuristic with the quasi-optimal policy (\textbf{C}) and Infotaxis (\textbf{D}). For this set of simulations, the side of the square lattice is $L=51$ with spacing $\Delta x=1$, the agent and target speed is $U=1$, the characteristic length scale of the likelihood of a detection is $\lambda=3$, the emission rate $R=1$, and the agent finds the source when $\|\bm{X}_{\rm ag}-\bm{X}_s\|\le\sqrt{2}$.}
    \label{fig_discr_RT}
\end{figure*}

In Fig.~\ref{fig_discr_RT}A (black curve), we show the average search times, obtained from the quasi-optimal strategy, as a function of the target's persistence time $\tau_p$. For relatively short $\tau_p$, the average capture time increases with $\tau_p$. However, for very long persistence times, the target tends to move approximately along a straight line, making it easier for the agent to intercept it. As a result, the average capture time curve features a maximum, hinting at the existence of an intermediate regime in which olfactory pursuit is the hardest.

In general, solving the POMDP is a challenging computational task, and only approximate solution methods are viable for relatively moderate-sized search arenas, furthermore requiring suitable fine-tuning of the numerical algorithm's hyperparameters. 
As an alternative heuristic baseline, we generalize Infotaxis~\cite{Vergassola2007} (see Methods) to the case of a moving source, a popular and successful strategy originally proposed for a fixed target, in which the agent moves to minimize the expected entropy of its belief.
The average search time obtained with Infotaxis is shown in Fig.~\ref{fig_discr_RT}A (blue curve). At small $\tau_p$, when the target diffuses very little from its initial position -- i.e., the target is approximately stationary -- Infotaxis is close to the quasi-optimal policy, only mildly underperforming it. However, at large $\tau_p$, Infotaxis fails to track the target and effectively reduces to a random search strategy (gray curve), where the agent performs a persistent random walk without using any olfactory cues (as described in Methods). As expected, the random searcher performs worse in the small-$\tau_p$ limit (target almost fixed), while it improves at large $\tau_p$, as a higher diffusivity of the source enhances the capture probability.

The deterioration of Infotaxis performance is mainly due to its bias toward exploration, making it ineffective when the target is motile with a large persistence time. {Indeed, as $\tau_p$ increases, the target's effective displacement over its run duration grows proportionally -- making its net speed increasingly comparable to that of the agent -- which defines a critical ``blind spot'' where purely exploratory strategies systematically fail to close the gap with the target.}
To mitigate this effect, we thus propose adding a greedy component that can balance excessive exploration when necessary, an approach pioneered in Ref.~\cite{Cassandra1996} (see also Refs.~\cite{masson2013olfactory,loisy2022}) in the context of a fixed target source. Such a greedy component is obtained by first solving the Bellman equation associated with the Markov decision process (MDP) under the assumption that the agent has perfect knowledge of the target state (position, velocity) and of the transition probability between states. 
The resulting recurrent relation can be solved by value iteration \cite{sutton1998reinforcement} for a range of $\tau_p$ values, thus generating 
the state-action value function $Q_{\textrm{MDP}}(\bm{X}_{\rm ag}-\bm{x},\bm{u};\bm{a})$, from which the optimal MDP policy follows (see Methods and Supplementary Fig.~S1). This state-value function is then averaged over the belief $b_t$ to obtain $Q_{\rm greedy}(\bm{a})=\langle  Q_{\textrm{MDP}}(\bm{X}_{\rm ag}-\bm{x},\bm{u};\bm{a})\rangle_b$.
Finally, we propose to blend exploitation and exploration by linearly combining greedy and infotactic contributions, so that the agent's action is selected according to the hybrid heuristic
\begin{equation}
    \bm{A}(t) = \argmax_{\bm{a}}  \left[w   Q_{\rm greedy}(\bm{a})  - (1-w)\mathcal{H}[b|\bm{a}] \right],
    \label{eq:blend}
\end{equation}
where $\mathcal{H}[b \mid \bm{a}]$ denotes the expected entropy of the belief after taking action $\bm{a}$, and $w \in [0,1]$ is a mixing parameter such that $w=0$ and $w=1$ correspond to purely infotactic and purely greedy strategies, respectively (see Methods and the pseudo-code in the Supplementary Material).
 As a result, we expect the agent to explore while the belief is spread and then bias its moves toward the target as the belief becomes sufficiently concentrated. The transition between these two limits is tuned by the weight $w$.

\begin{figure*}
    \centering
    \includegraphics[width=1.\linewidth]{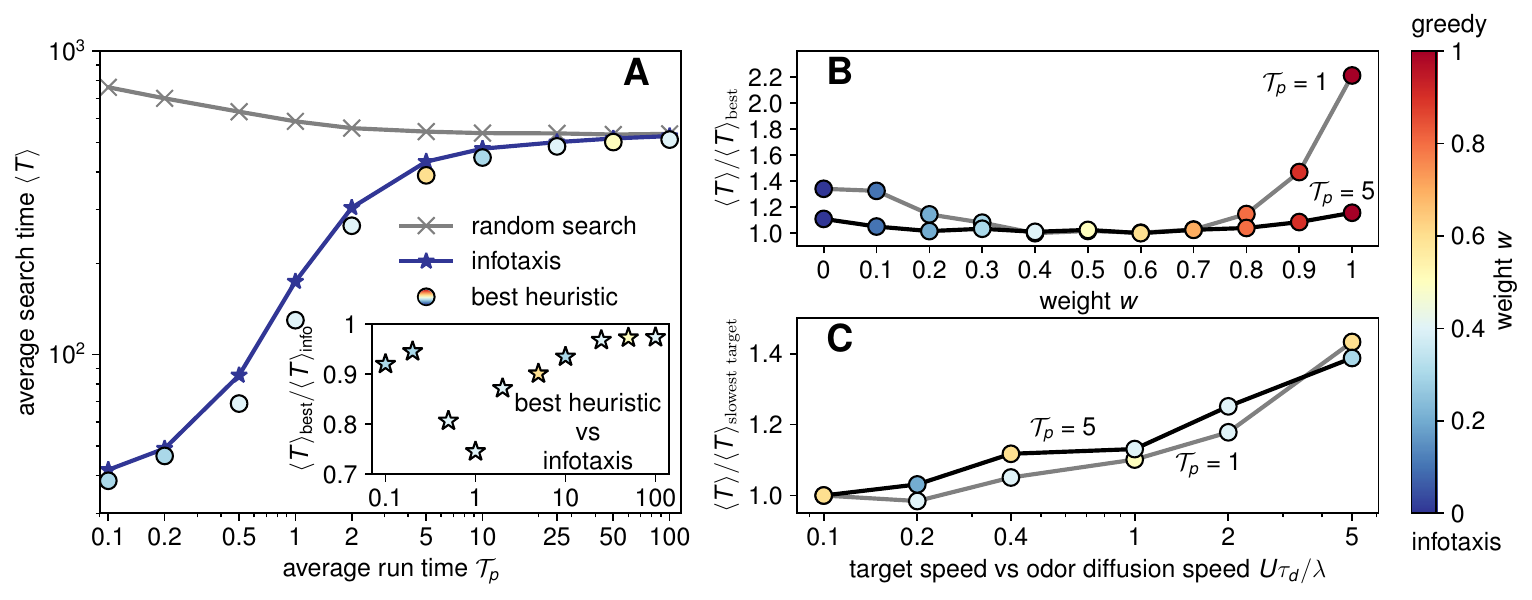}
    \caption{Results from the continuous run-and-tumble model, with the target performing a continuous run-and-tumble on a periodic square and the agent moving on a periodic lattice. Panel A shows the average search time $\langle T \rangle$ as a function of the target persistence time, averaged over $10^4$ search episodes starting from a detection: circles indicate the hybrid heuristic policy \ref{eq:blend} for the best weight $w$, color-coded as in the right bar; Infotaxis in blue corresponds to $w=0$;  the gray line refers to random search, obtained with the agent performing a persistent random walk without olfactory cues. The inset compares the average search times of the best hybrid heuristic with Infotaxis. Panel B shows the effect of the policy weight $w$ on the normalized average search time for two persistence times of the target motion. Panel \textbf{C} shows the normalized average search time as a function of $U\tau_d/\lambda$, i.e.~the ratio between the agent speed $U$ and the characteristic speed of the chemical diffusion $\lambda/\tau_d$, at emission rate $R=36/\tau_d$. Results in panels A and B correspond to $U\tau_d/\lambda=0.1$ at $R=120$. For the results reported here, the agent and target speed is $U=1$, the side of the lattice domain is $L=51$, the detection likelihood length scale is $\lambda=3$, and the agent finds the source when $\|\bm{X}_{\rm ag}-\bm{X}_s\|\le\sqrt{2}$.}
    \label{fig_cont_RT}
\end{figure*}

Remarkably, as shown by the colored circles in Fig.~\ref{fig_discr_RT}A, the hybrid heuristic policy closely follows the quasi-optimal solution and systematically outperforms Infotaxis, especially at large  $\tau_p$. This is further exemplified in the insets of Fig.~\ref{fig_discr_RT}A, showing the 
search time complementary cumulative distribution function (CCDF), i.e., the probability that the search time exceeds $T$.

For each  $\tau_p$, the circles in  Fig.~\ref{fig_discr_RT}A correspond to the strategy \ref{eq:blend} with the weight $w$ that yields the lowest average capture time $\langle T\rangle$. To this end, Fig.~\ref{fig_discr_RT}B shows that an optimal intermediate value of the weight exists. In particular, for large $\tau_p$, a pronounced minimum is reached for $w=0.9$ (i.e., a mostly greedy policy with just a small explorative infotactic component) with about $70\%$ improvement with respect to Infotaxis ($w=0$). This indicates the effectiveness of greedy strategies and highlights the importance of maintaining a degree of exploration. For smaller values of $\tau_p$, the optimal weight shifts towards lower values as the policy's greediness becomes less critical, consistently achieving performance closer to (but yet better than) Infotaxis. 
Finally, Fig.~\ref{fig_discr_RT}C shows that the best hybrid [ achieves performance close to the quasi-optimal policy over a wide range of target persistence times, with $\langle T\rangle_{\rm best}/\langle T\rangle_{\rm POMDP}$ remaining near unity and noticeable deviations arising only around $\tau_p\approx 5$. Comparison with Infotaxis (Fig.~\ref{fig_discr_RT}D) reveals a stronger dependence on persistence time, indicating a crossover between regimes where information-driven exploration is effective (small $\tau_p$) and those in which exploitation is more advantageous (large $\tau_p$). 

The Supplementary movies S1-S2 in the Supplementary Material show sample trajectories for the best hybrid heuristic strategy at different $\tau_p$: at a short persistence time, the agent exhibits characteristic infotactic spirals, while at larger $\tau_p$ it transitions to direct chasing behavior. Spirals are associated with exploratory behavior and are very effective when the target moves little, whereas direct-chasing trajectories are associated with the need to pursue a target that moves persistently. Movies S3-S4 compare sample realizations from the best hybrid heuristics and Infotaxis at $\tau_p=2$ and $25$, displaying the more exploitative trend of the best hybrid heuristic.

Overall, the heuristic policy provides a near-optimal yet computationally efficient alternative to solving the POMDP and adapts more robustly to changes in the target persistence time than Infotaxis. 
These findings support the use of the heuristic policy to tackle more complex \sout{and realistic} scenarios, where the POMDP solution cannot be effectively approximated.

\subsection*{Continuous run-and-tumble}

We next analyze a more complex setup, in which the target performs a continuous run-and-tumble (R\&T) motion. This setup is inspired by bacterial motion \cite{berg2025random} and the alternation of straight-line and zigzag evasion of a prey from a predator \cite{humphries1970protean,furuichi2002dynamics}. In the model, the target moves off-lattice and in continuous time according to the dynamics
\begin{equation}
\dot{\bm X}_s = U \bm n(\phi); \qquad
\dot{\phi} = \sum_{n=0}^\infty \phi_n \delta(t-t_n) 
\label{eq:RTc}
\end{equation}
where $\bm n(\phi)=(\cos\phi,\sin\phi)$ denotes the direction of motion. The tumbling angle $\phi_n$ is random uniform  over $[0,2\pi)$, while the interval between tumbling times $t_{n+1}-t_{n}$ is exponentially distributed with mean $\mathcal{T}_p$. 

During movement, the target emits olfactory signals -- particle tracers -- that undergo Brownian motion and have exponentially distributed lifetimes with mean $\tau_d$. 
We assume that the agent, which moves at discrete times, with step $\Delta t=1$ (corresponding to the sensing/decision time), detects an olfactory cue when at least one odor particle enters a small (sensing) region around it. However, unlike the discrete setup, we simulate an ensemble of odor tracers, following their trajectories from the target position and diffusing until their lifetimes. This setup thus accounts for the persistence of odor particles emitted by the target, thereby generating a trail behind it, a feature absent in the previous model. As a consequence, if the target moves faster than the characteristic diffusion speed of odor tracers, the detected odor particles may not fully represent the target position, thereby complicating the agent's task. 

To keep the search realistic, we let the agent use an  approximated model of cue detection, obtained by assuming instantaneous diffusion of the tracers, leading to the likelihood of detection \ref{eq:likeRT}, discussed in Methods.
As in the previous setup, the agent moves on a discrete lattice by choosing one of its four neighbor sites. Furthermore, it has a discretized (and thus inaccurate) model of the target dynamics in which the target can jump to the four nearest neighbors, the four diagonal neighbors, or remain stationary. The model is encoded in a transition matrix empirically obtained by discretizing on a lattice the trajectories generated from long-time integration of Eq.~\ref{eq:RTc}, as outlined in Methods (see also Fig.~S2 in SM).

In this more {complex} framework, where the POMDP approach is inapplicable due to the need to use an approximate model of the target's dynamics, the use of an effective heuristic strategy is further motivated \cite{heinonen2023,heinonen2025b}.
We therefore employ the same strategy \ref{eq:blend} proposed in the discrete setup, selecting the weight that yields the minimum average capture time $\langle T\rangle_{\rm best}$, and compare its performance with Infotaxis, based on the same approximated model of the target dynamics. 
As reported in Fig.~\ref{fig_cont_RT}A, for not too large persistence times, the heuristic yields an improvement of approximately $25\%$ over Infotaxis (see inset, and also Supplementary movie S5 for an example of the agent and target dynamics with $\mathcal{T}_p=1$), while the gain saturates at large $\mathcal{T}_p$, where both Infotaxis and the best hybrid heuristic tend to the random search limit (gray curve).
The poor performances at large $\mathcal{T}_p$ are mainly due to errors in the agent's model of the target motion, which hinders the correct identification of the target persistent trajectory, under the one-step Markov approximation of the dynamics. 
In fact, even when the target moves with perfectly persistent motion, the discretization causes the agent to interpret its motion as having only finite persistence.
Building a more accurate model of the discretized dynamics, e.g., by using a multi-step Markov process, could boost the performance of the heuristic policy for large persistence times. However, this goes beyond the main scope of this paper.

As shown in Fig.~\ref{fig_cont_RT}B, the normalized average capture time has a minimum as a function of the blending weight $w$ for $w\approx 0.5$, around which it remains relatively flat, indicating that there is no need to fine-tune this parameter. To assess the robustness of the policy to errors in the detection model, we also analyze the average search time as a function of the target's speed, compared to the characteristic speed of diffusion of odor particles, encoded in the parameter $U\tau_d/\lambda$, while adjusting the emission rate $R$ so that the average number of odor particles, $R\tau_d$, in the domain remains constant. 
Figure~\ref{fig_cont_RT}C shows that for not too large values of this normalized target's speed $U\tau_d/\lambda$, the average capture time does not deteriorate significantly, indicating the heuristic policy's robustness to errors in the detection model.
Moreover, although the mean search time grows with $U\tau_d/\lambda$, the relative gain of the best hybrid heuristic over Infotaxis remains essentially constant (not shown).

The continuous run-and-tumble setting highlights the challenges posed by {highly complex}  target motion and imperfect internal models. Despite significant modeling errors and the breakdown of the POMDP approach, the proposed heuristic policy consistently outperforms Infotaxis, especially at intermediate persistence times. The weak dependence of performance on the blending weight and the target's characteristic speed relative to the odor tracers' diffusion speed further demonstrates the robustness of the strategy. These results indicate that the combination of exploration-driven and greedy components provides a reliable and flexible framework for olfactory pursuit in complex environments.

\section*{Discussion}

This study addresses the problem of tracking a moving source in a noisy environment using olfactory cues. We demonstrate that while information-theoretic approaches like Infotaxis are effective for stationary or low-persistence targets, they are insufficient for tracking prey exhibiting high directional persistence. In such regimes, the optimal strategy shifts toward predictive planning, in which the agent uses a model of the prey's motion to anticipate its future position.

Our results address a critical ``blind spot'' where the predator and prey move at comparable speeds. In this domain, neither pure exploration (Infotaxis) nor pure exploitation (greedy planning) is sufficient. Instead, a hybrid heuristic that dynamically balances reducing belief entropy with maximizing expected value proves to be the most robust solution. This composite policy allows the agent to adapt to the target's behavior, prioritizing information gathering when the prey's path is uncertain and switching to interception when the trajectory becomes predictable. We validated this approach in a continuous run-and-tumble model, confirming that the heuristic remains effective even when the agent must contend with the unsteady nature of odor wakes that lag behind the source.

The present work opens several avenues for future research.
First, while we assumed that the agent has exact prior knowledge of the 
target's motion statistics and environmental parameters, such as diffusivity 
and decay rates, these quantities are rarely available in biological or 
robotic scenarios. A natural extension would incorporate online or 
reinforcement learning, enabling the agent to estimate the transition matrix 
of the prey and the relevant properties of the olfactory environment 
in real time.

Second, the environmental model relies on \sout{simplified} diffusion and linearized approximations of the concentration field, which may not fully capture the complex intermittency and correlations found in high-Reynolds-number turbulence. Future work should test these strategies in fully turbulent fluid simulations or experimental settings to assess their robustness to structurally complex odor plumes.

Finally, the current study treats the source as a non-adversarial entity that moves independently of the searcher. Extending this framework to a game-theoretic setting, in which the prey actively detects and evades the predator, would introduce co-evolutionary dynamics likely to require even more sophisticated, and likely stochastic, pursuit and evasion strategies {\cite{borra2022}}. Additionally, scaling the search domain from two to three dimensions would increase the computational complexity of belief updates, potentially requiring more efficient approximation methods than the ones employed here.

\matmethods{

\subsection*{Discrete run-and-tumble model}

In the discrete run-and-tumble (R\&T) model, the target position evolves discretely in time with a time step $\Delta t=1$, according to $\bm{X}_s(t+1)= \bm{X}_s(t) + \bm{U}_s(t)$ with the velocity governed by the Markov chain
\begin{align}
\bm{U}_s(t+1)=\left\{
\begin{array}{ll}
\bm{U}_s(t) &\textrm{with prob. } \, \epsilon\phantom{/4}  \textrm{ if } \bm{U}_s(t)\ne\bm{0} \\ 
\textrm{any } \bm{u}\ne\bm{0} &\textrm{with prob. } \, \epsilon/4  \textrm{ if } \bm{U}_s(t)=\bm{0} \\ 
\bm 0 &\textrm{with prob. } \, 1-\epsilon
\end{array} \right.
\label{eq:discrete}
\end{align}
where $\bm{u}\in \{\bm{e}_1,-\bm{e}_1, \bm{e}_2,-\bm{e}_2, \bm{0}\}$ and $\bm{e}_i$ are the unit vectors of the standard basis. Thus, the target moves on a lattice with spacing $\Delta x =1$. 

The transition matrix $P(\bm{u}'|\bm{u})$, namely the probability of the velocity at the next step $\bm{u}'=\bm{U}_s(t+1)$ given the current velocity $\bm{u}=\bm{U}_s(t)$, associated with Eq.~\ref{eq:discrete} features the invariant probability $q(\bm u)$
which is equal to $\epsilon/4$ for velocities along the cardinal directions and to $1-\epsilon$ for the rest state.

The characteristic persistence times are $\tau_p = 1/(1-\epsilon)$ for a state of motion and $\tau_0 = 1/\epsilon$ for the resting state $\bm{0}$, recovering a fixed target in the limit $\epsilon \to 0$.

\subsection*{Model of the Environment}

During its motion, the target emits odor particles that diffuse and decay in the underlying fluid, resulting in the concentration field $\theta(\bm{x},t)$ advected by a turbulent flow $\bm{v}(\bm{x},t)$,
\begin{equation} 
    \partial_t \theta + \bm{v}\cdot\nabla \theta = \overline{\kappa}\nabla^2 \theta - \frac{\theta}{\tau_d} + R\delta(\bm{x}-\bm{X}_s(t)),
    \label{eq:conc}
\end{equation}
where $R$ is the emission rate that we set to 1, $\bm{X}_s$ the source position, $\tau_d$ the odor decay timescale, and $\overline{\kappa}$ the molecular diffusivity.

Averaging over velocity realizations and approximating turbulent transport by an effective advection–diffusion process yields an equation for the mean concentration $\langle\theta\rangle(\bm{x},t|\bm{X}_s(t))$,
\begin{equation}
    \partial_t \langle\theta\rangle + \bm{V}\cdot\nabla \langle\theta\rangle = \kappa \nabla^2 \langle\theta\rangle - \frac{\langle\theta\rangle}{\tau_d} + R \delta(\bm{x}-\bm{X}_s),
    \label{eq:conc_avg}
\end{equation}
where $\bm{V}=\langle\bm{v}\rangle$ is the mean wind and $\kappa=\overline{\kappa}+\kappa_T$ includes both molecular and turbulent diffusion.

To compute the likelihood of a detection used in the agent's model of the environment, we need to solve Eq.~\ref{eq:conc_avg} for the mean concentration $\langle\theta\rangle(\bm{x},t|\bm{X}_s(t))$. We approximate the solution 
assuming steady state in Eq.~\ref{eq:conc_avg}, in the absence of a mean wind, for a fixed source, evaluated at the instantaneous source position, which yields
\begin{equation}
    \langle\theta\rangle(\bm{x}|\bm{X}_s(t)) = \frac{R\tau_d}{2\pi\lambda^2} K_0\left(\frac{\|\bm{x}-\bm{X}_s(t)\|}{\lambda}\right),
    \label{avg_conc}
\end{equation}
where $K_0$ is the modified Bessel function of order zero and $\lambda=\sqrt{\kappa\tau_d}$ is the characteristic decay length.
Then, using standard Smoluchowski arguments \citep{smoluchowski,Vergassola2007,loisy2022}, the expected mean number of detected particles within a time interval $\Delta t$ depends only on the distance from the source  and reads
\begin{equation}
    \mu(\bm{x}-\bm X_s(t)) = \frac{2\pi \kappa \Delta t }{\ln(2\lambda/\Delta x)}\langle \theta\rangle (\bm{x}|\bm{X}_s(t))
    \label{avg_conc}
\end{equation}
 This quasi-static approximation is valid provided that the source displacement during the decay time scale is small compared to $\lambda$, namely
\begin{equation}
    \tau_d U \ll \lambda,
    \label{steady_approx}
\end{equation}
with $U$ the target speed.

Binary odor detections, $H(t)\in\{\texttt{no-detection},\texttt{detection}\}$ are generated via a Bernoulli process with
probabilities
\begin{subequations}
\begin{align}
\mathcal{L}(\texttt{no-detection}|\bm{X}_{\rm ag}-\bm{X}_s) &\equiv \exp(-\mu(\bm{X}_{\rm ag}-\bm{X}_s))  \label{eq:l0}\\
\mathcal{L}(\texttt{detection}|\bm{X}_{\rm ag}-\bm{X}_s)  &\equiv 1-\exp(-\mu(\bm{X}_{\rm ag}-\bm{X}_s))   \label{eq:l1} 
\end{align}
\label{eq:likelihood01}
\end{subequations}
where $\bm{X}_{\rm ag}$ is the agent position and $\mathcal{L}(H(t)|\bm{X}_{\rm ag}-\bm{X}_s)$ denotes the detection likelihood.

\subsubsection*{Olfactory cues in the discrete R\&T}
In the discrete R\&T setup, we assume that the concentration field is in a steady state and that Eq.~\ref{steady_approx} holds. Under these hypotheses, relaxed in the continuous R\&T scenario, the concentration field of the chemical signal emitted by the target is given by Eq.~\ref{avg_conc}, with a characteristic length scale $\lambda=3$. In this setup, the model of the agent for the likelihood of a detection is exact.

\subsubsection*{Olfactory cues in the continuous R\&T}
Here, differently from the discrete run-and-tumble case, we simulate the odor particles adopting a Lagrangian viewpoint on Eq.~\ref{eq:conc}. This allows us to account for the unsteadiness of the trail of tracers, which forms behind the target and steers due to its motion. While moving, the source emits at each time step a Poissonian number of particles with mean $N_{e}=R\mathrm{d}t$. Odor particles undergo a Brownian motion, $\dot{\bm X}_i =\sqrt{2\kappa}\bm{\eta}$ where $\bm{\eta}$ zero-mean, white-in-time Gaussian noise with isotropic correlation, $\langle\bm{\eta} \bm{\eta}^\top(t')\rangle = \bm{I}\delta(t-t')$. The tracers, emitted at time $t_i$, start at the source position, $\bm{X}_i(t_i) = \bm{X}_s(t_i)$, and diffuse for a time $\tau_i$  exponentially distributed with mean $\tau_d$. 

The agent detects an odor signal if one or more tracers are at a distance below a detection radius, set equal to the spacing $\Delta x$ of the lattice on which the agent moves. In this { physics-based}  description of the odor unsteady landscape, the agent uses an approximate likelihood of a detection, resulting in an incorrect model of the environment. We do not define the mean hit rate using the Smoluchowski equation \cite{Vergassola2007}. Instead, we use the probability that a particle is in a cell of side $\Delta x$ centered at $\bm{X}_{\rm ag}$ after being emitted from a fixed source located at $\bm{X}_s$, resulting in the likelihood of a detection 
\begin{equation}
\overline{\mathcal{L}}(\texttt{detection}|\bm{X}_{\rm ag}-\bm{X}_s) = 1-\exp(-\Delta x^2 \langle\theta\rangle(\bm{X}_{\rm ag}|\bm{X}_s(t))) \,, \label{eq:likeRT}
\end{equation}
which is accurate only if Eq.~\ref{steady_approx} holds. 

\subsection*{Bayes update of the belief}

The search is initialized at $t=0$ with a detection: the target position is drawn from the likelihood \ref{eq:likelihood}, set to zero in the surrounding of the agent (where it would have already found the target) and properly renormalized, while its initial velocity is sampled from the invariant distribution $q(\bm u)$ of the discrete model. In the continuous case, the initial velocity is sampled from the invariant probability obtained from the discretized Markov model. Consequently, the initial belief is given by the product of the likelihood and the stationary probability of the target velocity.

At each discrete time $t$, the agent’s information about the target state (position and velocity), encoded in the belief $b_t(\bm{x},\bm{u})$, is updated based on the known target dynamics through the transition probability $P(\bm{u}'|\bm{u})$:
\begin{equation}
    b'_{t}(\bm{x}',\bm{u}') = \sum_{\bm{u}} P(\bm{u}'| \bm{u}) b_{t}(\bm{x}'-\bm{u},\bm{u})\,.    
\end{equation}
This propagates the belief, producing a probability map of the target’s new state conditioned on the previous belief. At this point, the agent takes its action; if it does not find the source, it continues to update its belief as follows. First, it sets to zero the belief around its position, at lattice points within the capture radius (i.e., when $\|\bm{x}-\bm{X}_{\rm ag}\| \le \sqrt{2}$), and properly renormalizes it:
\begin{equation}
\label{eq:normalize}
 b'_{t+1}(\bm{x},\bm{u}) = \begin{cases}
 0 \textrm{ if } \|\bm{x}-\bm{X}_{\rm ag}\| \le \sqrt{2}\\
 b'_{t}(\bm{x},\bm{u})/(1-p_f) \textrm{ otherwise}
 \end{cases}
\end{equation}
where $p_f = \sum_{\bm{u},\|\bm{x}-\bm{X}_{\rm ag}\| \le \sqrt{2}} b'_t(\bm{x},\bm{u})$ is the expected probability to having found the source.
Then, finally, it updates the belief using the odor detection $H(t)$ at its position $\bm{X}_{\rm ag}(t)$ via Bayes’ rule:
\begin{equation}
\label{eq:bayes}
 b_{t+1}(\bm{x},\bm{u}) = \frac{\mathcal{L}(H(t)|\bm{X}_{\rm ag}(t)-\bm{x}) b'_{t+1}(\bm{x},\bm{u})}{p(H(t))}
\end{equation}
where the expected probability of measuring $H(t)$, $p(H(t))$, follows from the normalization $\sum_{\bm{x},\bm{u}} b_{t+1}(\bm{x},\bm{u})=1$. A detailed step-by-step description of the olfactory pursuit algorithm is given in SM.

\subsection*{POMDP} We find a quasi-optimal policy similarly as in Refs.~\cite{loisy2022,heinonen2023,loisy2023}.
At each time step, we have three possibilities: the source is found if $|\bm{X}_{\rm ag}-\bm{X}_{s}| \le \sqrt{2}$, or otherwise the agent observes either $\texttt{detection}$ or $\texttt{no-detection}$. The probabilities of making these three mutually exclusive observations are given, respectively, by the probability of the agent being within a distance $\sqrt{2}$ from the source, and by Eq.~\ref{eq:likelihood01}.
Also, the transition probability between states 
$P(\bm{X}_{\textrm{ag}}'-\bm{x}',\bm{u}'|\bm{X}_{\textrm{ag}}-\bm{x},\bm{u},\bm{a})$ follows from the transition matrix describing the target motion $P(\bm{u}' |\bm{u})$ given by Eq.~\ref{eq:discrete} and the dynamical evolution rules for the agent and the target. 
The reward is unity upon arrival at the source and zero otherwise. Finally, as for the heuristic policy, we initialize searches with a detection, so that the prior $b_0(\bm{x}, \bm{u})$ is proportional to $\mathcal{L}(\texttt{detection}|\bm{X}_{\rm ag}-\bm{x})q(\bm{u})$.

The definitions above suffice to specify the POMDP. The belief-state value function $V[b]$ is then the unique solution to the Bellman equation \ref{eq_bellman}~\cite{astrom1965,sondik1978,kaelbling1998}. The solution of this equation is computationally intractable, but approximate methods are viable when the state space is not too large (e.g., $O(10^4)$ elements). We use the SARSOP algorithm \cite{sarsop} to represent $V$ as a set of vectors $\bm{\alpha}$ such that $V[b] = \max_{\bm{\alpha}} \bm{\alpha} \cdot \bm{b}$ (with $\cdot$ a product and summation over states). The quasi-optimal policy is then given by $\pi(b) =\argmax_{\bm{a}} Q[b,\bm{a}]$. In practice, we take the discount factor $\gamma$ as close as possible to unity, in which limit the objective function becomes the expected arrival time to the source (as $\gamma \to 1,$ the POMDP planning problem becomes exponentially more complex). For each $\tau_p$, we obtain quasi-optimal policies for both $\gamma=0.98,0.99$ and select the policy that yields a lower mean empirical arrival time on an ensemble of search trials.

\subsection*{Heuristic policies}

\subsubsection*{Infotaxis}
Infotaxis was introduced in the framework of olfactory search of a fixed source \cite{Vergassola2007}, and it can be extended to olfactory pursuit of a moving source. The basic idea is to formulate the problem as a sequential Bayesian inference problem, in which the agent maintains a belief $b(\bm{x},\bm{u})$ over the target state and chooses its actions by minimizing the expected uncertainty in the target state. Such uncertainty is quantified by the Shannon entropy of the belief,
\begin{equation}
\mathcal{H}[b] = - \sum_{\bm{x},\bm{u}} b(\bm{x},\bm{u})\,\log_2 b(\bm{x},\bm{u}) \, .
\end{equation}
At time $t$, the infotactic agent selects the action $\bm{A}(t)$ that is expected to maximize the reduction of the belief entropy 
\begin{equation}
    \bm{A}(t) = \arg\max_{\bm{a}} \left( \mathcal{H}[b] - \mathcal{H}[b|\bm{a}]\right)
\end{equation}
which is equivalent to minimizing $\mathcal{H}[b|\bm{a}]$ \cite{PanizonCelani}. 
This strategy promotes exploratory moves when the belief is broad and progressively concentrates on high-probability regions as the posterior sharpens.

\subsubsection*{Greedy (MDP-based) component}
To incorporate exploitation in the agent's strategy, we combine Infotaxis with the policy obtained by solving a fully observable Markov Decision Process (MDP) in which the target state $(\bm{X}_s,\bm{U}_s)$ is assumed known. In particular, we solve the Bellman equation \ref{eq_bellman} via value iteration \cite{sutton1998reinforcement}, with discount factor $\gamma=0.95$, for a belief collapsed on the actual state of the target, $b_t(\bm{x},\bm{u})=\delta(\bm{x}-\bm{X}_s(t))\delta(\bm{u}-\bm{U}_s(t))$.  
The resulting state-action matrix $Q_{\rm MDP}(\bm{X}_{\rm ag}-\bm{x},\bm{u};\bm{a})$ is shown in Supplementary Fig.~S1 for $\tau_p=2$ and 25. At short persistence times, the greedy policy coincides with moving towards the target following the shortest path. However, for a long persistence time, ambush actions are more convenient because the agent can confidently predict the target's trajectory over extended time horizons.

When the agent does not exactly know the state of the target, and uses odor cues to infer it, the state-action matrix from the solution of the MDP is averaged over the current belief $b_t(\bm{x},\bm{u})$, to get the greedy part of the heuristic policy
\begin{eqnarray}
    Q_{\rm greedy}(\bm{a}) &=& \langle Q_{\mathrm{MDP}}(\bm{X}_{\rm ag}-\bm{x},\bm{u};\bm{a}) \rangle_{b} = \nonumber\\
    &=& \sum_{\bm{x},\bm{u}} b(\bm{x},\bm{u})\, Q_{\mathrm{MDP}}(\bm{X}_{\rm ag}-\bm{x},\bm{u};\bm{a}).    
\end{eqnarray}

\subsubsection*{Hybrid heuristic policy}
The heuristic strategy combines information gain and exploitation by linearly blending the two objectives as indicated in Eq.~\ref{eq:blend}, where the blending weight $w \in [0,1]$ controls the exploration--exploitation trade-off. 
The limits $w=0$ and $w=1$ recover Infotaxis and pure greedy pursuit, respectively. 
For each parameter set (e.g., persistence time), $w$ is selected to minimize the average capture time computed over simulated search episodes.
This construction preserves the information-seeking behavior required when the belief is diffuse, while enabling anticipatory interception when the posterior over the target's position and velocity becomes sufficiently concentrated.

\subsection*{Discrete approximation for the continuous run-and-tumble model}
In order to build a discrete Markov model for the dynamics \ref{eq:RTc}, we proceed as follows. Equation~\ref{eq:RTc} is integrated for intervals of time $\Delta t=1$, and the positions $\bm X_s(t)$ are mapped to the discrete lattice where the agent moves: $\hat{\bm X}_{s}(t)=[\bm X_s(t)/\Delta x]$ (where $\Delta x=1$ is the lattice spacing and $[\cdot]$ denotes the integer part). We can then define the discrete velocity $\bm{U}_s(t)=\hat{\bm X}_{s}(t+1)-\hat{\bm X}_{s}(t)$.
Because the target speed is fixed to $U=1$, this discretized velocity can take only nine possible values, corresponding to motion along the four cardinal directions, the four diagonals, or no displacement.
Finally, using a long simulated trajectory, we estimate the transition probabilities $P(\bm{U}_s(t+1)\mid \bm{U}_s(t))$ numerically. In doing so, we also enforce the symmetry properties of the transition matrix (see Supplementary Fig.~S2 for details).

\subsection*{Random search without olfactory cues}
In the absence of olfactory cues, the best the agent can do is a random search. Inspired by Ref.~\cite{benichou2012}, we modeled this as a simple persistent random walk, given by a variant of Eq.~\ref{eq:discrete}:
\begin{align}
    \bm a(t+1)=\left\{
    \begin{array}{ll}
        \bm a(t) &\textrm{with prob. } \, 1-\alpha  \\ 
        \textrm{any } \bm{a}\ne\bm{0} &\textrm{with prob. } \, \alpha \,, \\ 
    \end{array} \right.
    \label{eq:random-search}
\end{align}
where, as in the olfactory search, the agent’s actions $\bm{a}$ are restricted to the cardinal directions.
The average search time was computed numerically, with agent and target initial positions drawn from the same distribution used in the olfactory search. The persistence parameter $\alpha$ was scanned to identify the value minimizing the mean search time, which was then used as a benchmark for the olfactory-based strategies, as shown in Figs.~\ref{fig_discr_RT} and~\ref{fig_cont_RT}.

}

\showmatmethods{}

\acknow{We acknowledge useful discussions with Massimo Vergassola. 
We acknowledge financial support under the National Recovery and Resilience Plan (NRRP),
Mission 4, Component 2, Investment 1.1, Call for tender No.\ 104 published on 2.2.2022 by the Italian Ministry of University and Research (MUR), funded by the European Union — NextGenerationEU — Project Title Equations informed and data-driven approaches for collective optimal search in complex flows (CO-SEARCH), Contract 202249Z89M — CUP B53-D23003920006 and E53-D23001610006.
This work was also supported by the Italian Ministry of University and Research (MUR) - Fondo Italiano per la Scienza (FIS2) - 2023 Call, project DeepFL, CUP E53C24003760001, and by the European Research Council (ERC) under the European Union’s Horizon 2020 research and innovation program (Grant Agreement Nos.\ 882340 and 101002724).}

\showacknow{} 





\bibsplit[15]

\bibliography{pnas-sample}



\section*{Supplementary material}

# Supplementary material for "Olfactory pursuit: catching a moving odor source in complex flows"

Maurizio Carbone[a,b] Lorenzo Piro[b], Robin A. Heinonen[c], Luca Biferale[b], Massimo Cencini[a], and Antonio Celani[d,e]

[a] Istituto dei Sistemi Complessi, CNR-ISC, Via dei Taurini 19, 00185 Rome,
Italy and INFN "Tor Vergata", Via della Ricerca Scientifica 1, 00133 Rome, Italy

[b] Department of Physics & INFN, Tor Vergata University of Rome,
Via della Ricerca Scientifica 1, 00133 Rome, Italy

[c] Machine Learning Genoa Center (MaLGa) & Department of Civil,
Chemical and Environmental Engineering, University of Genoa, Genoa, Italy

[d] Quantitative Life Sciences, The Abdus Salam International Centre for Theoretical Physics, 34151 Trieste, Italy and

[e] Department of Oncology, University of Torino, Torino, Italy

## I. SUPPLEMENTARY FIGURES

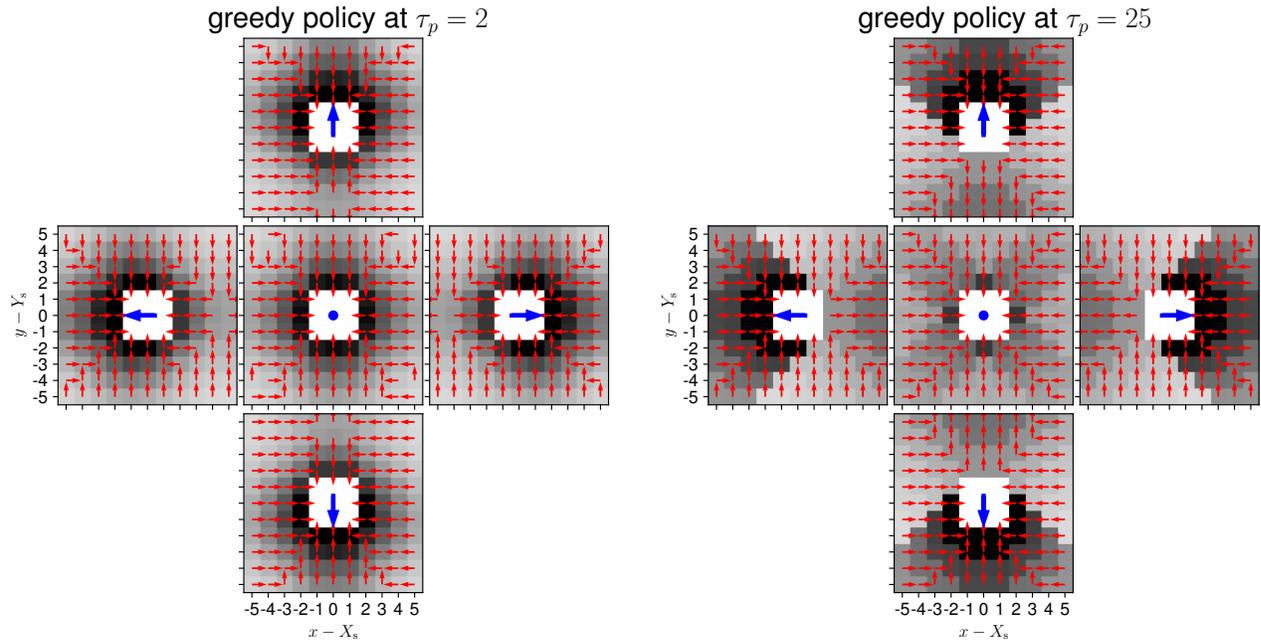

FIG. S1: Sketch of the greedy policy from the solution of the MDP on an $11 \times 11$ grid, with a target performing a discrete run-and-tumble. The blue arrow at the center indicates the action of the target. For each action of the target, the colormap in grayscale shows the value function (darker indicates a higher value) as a function of the separation between the agent and the target. The red arrows indicate the optimal action for each state.

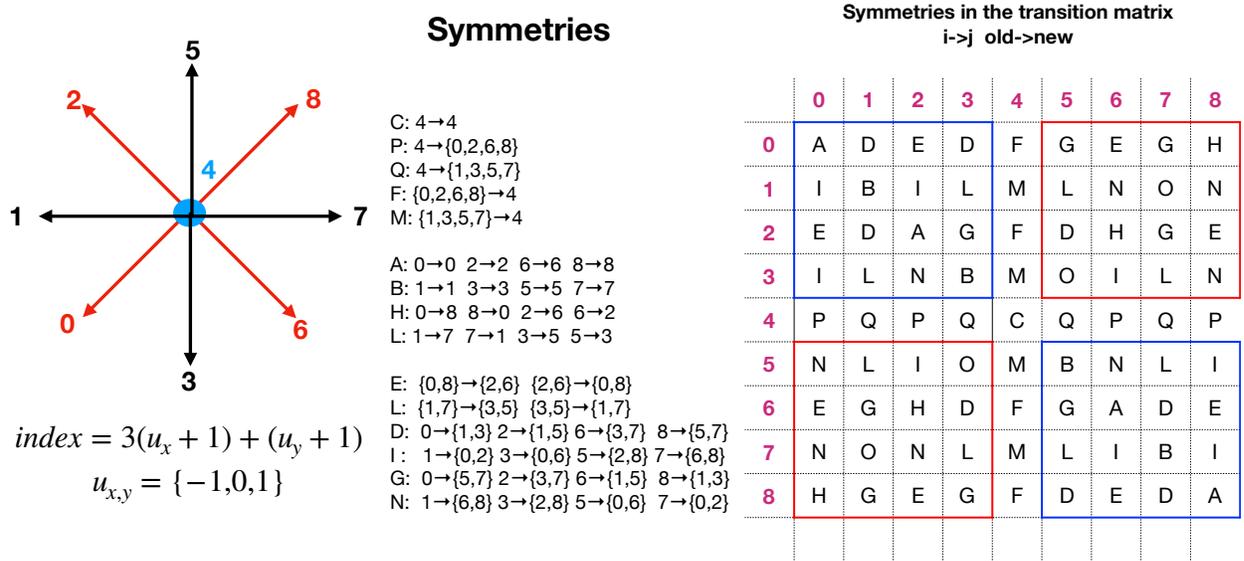

FIG. S2: Discrete Markov model for the continuous random and tumble model Eq.(4) of main text: given the strategy discussed in Methods and the target velocity modulus being $U = 1$, in a time step $\Delta t = 1$ for any position of the target only nine possible velocities are allowed $(u_x, u_y) = (-1, 0, 1, -1, 0, 1)$, corresponding to those shown on the left panel that can be indexed as described on the bottom left. Therefore, in principle, the transition probability matrix has $81 = 9 \times 9$ entries. However, using the symmetries of the problem, many transitions are constrained to be equal (see the middle panel), so that only 15 entries of the matrix are independent. The resulting (schematic) matrix is represented on the right panel. Such symmetries are used while numerically estimating the transition matrix. Note that the resulting discrete model only approximates the continuous dynamics and does not allow the agent to accurately infer the true persistence of the target motion. For example, consider a target starting at $(0.5, 0.5)$ and moving indefinitely along the diagonal at angle $\pi/4$ with respect to the horizontal axis. After projection onto the lattice, the corresponding discrete velocities would be $\boldsymbol{u}(0) = (1, 1)$, $\boldsymbol{u}(1) = (0, 0)$, $\boldsymbol{u}(2) = (1, 1)$, $\boldsymbol{u}(3) = (1, 1)$, $\boldsymbol{u}(4) = (1, 1)$, $\boldsymbol{u}(5) = (0, 0)$, $\boldsymbol{u}(6) = (1, 1)$, $\boldsymbol{u}(7) = (1, 1)$, $\boldsymbol{u}(8) = (0, 0)$, .... The inferred transition matrix thus fails to encode the underlying perfectly persistent motion in a fixed direction. As a consequence, this approximation becomes most pronounced for large persistence times $\mathcal{T}_p$.

**Algorithm 1** Olfactory pursuit: the input comprises the transition matrix $P$ describing the target motion, the invariant probability $q$ of the target velocity, the likelihood of a detection $\mathcal{L}$, and the heuristic policy blending weight $w$.

1: **Initialize (t=0)**
2: $\boldsymbol{X}_{\text{ag}}(0) = \boldsymbol{0}$ ▷ Set agent position
3: $b_0(\boldsymbol{x}, \boldsymbol{u}) \propto \mathcal{L}(\texttt{detection}|\boldsymbol{X}_{\text{ag}}(0) - \boldsymbol{x})q(\boldsymbol{u})$ ▷ Set initial normalized belief
4: $(b_0(\cdot), p_f) = \textsc{renomalize\_belief}(b_0, \boldsymbol{X}_{\text{ag}}(0))$
5: $(\boldsymbol{X}_s(0), \boldsymbol{U}_s(0)) \sim b_0(\boldsymbol{x}, \boldsymbol{u})$ ▷ Sample source initial condition

6: **for** each time step $t \leq T_{max}$ **do**
7: $\quad (\boldsymbol{X}_s(t+1), \boldsymbol{U}_s(t+1)) = \textsc{move\_source}(\boldsymbol{X}_s(t), \boldsymbol{U}_s(t), P)$
8: $\quad b_t'(\cdot) = \textsc{transport\_belief}(b_t, P)$ ▷ Transport belief
9: $\quad$ **for** each candidate action $\boldsymbol{a}$ **do**
10: $\quad\quad F(\boldsymbol{a}) = \textsc{compute\_heuristic}(b_t, b_t', \boldsymbol{X}_{\text{ag}}(t), \boldsymbol{a})$
11: $\quad$ **end for**
12: $\quad \boldsymbol{A}(t) = \arg\max_{\boldsymbol{a}} F(\boldsymbol{a})$ ▷ Select optimal action
13: $\quad \boldsymbol{X}_{\text{ag}}(t+1) = \boldsymbol{X}_{\text{ag}}(t) + \boldsymbol{A}(t)$ ▷ Agent moves
14: $\quad$ **if** $\|\boldsymbol{X}_{\text{ag}}(t+1) - \boldsymbol{X}_s(t+1)\| \leq \sqrt{2}$ **then**
15: $\quad\quad$ Return **source found** ▷ Search episode ends
16: $\quad$ **else**
17: $\quad\quad H(t+1) \sim \mathcal{L}(h|\boldsymbol{X}_{\text{ag}}(t+1) - \boldsymbol{X}_s(t+1))$ ▷ Sample odor signal from likelihood of a detection
18: $\quad\quad (b_{t+1}'(\cdot), p_f) = \textsc{renormalize\_belief}(b_t', \boldsymbol{X}_{\text{ag}}(t+1))$
19: $\quad\quad (b_{t+1}(\cdot), p_h) = \textsc{update\_belief}(b_{t+1}', \boldsymbol{X}_{\text{ag}}(t+1), H(t+1))$ ▷ Update belief based on odor detection
20: $\quad$ **end if**
21: **end for**
22: **end main**

23: **function** $\textsc{renormalize\_belief}(b, \boldsymbol{X}_{\text{ag}})$
24: $\quad p_f = \sum_{\|\boldsymbol{r}\| \leq \sqrt{2}, \boldsymbol{u}} b(\boldsymbol{X}_{\text{ag}} + \boldsymbol{r}, \boldsymbol{u})$ ▷ Probability of finding the source
25: $\quad b(\boldsymbol{x}, \boldsymbol{u}) \leftarrow \begin{cases} b(\boldsymbol{x}, \boldsymbol{u})/(1 - p_f) & \text{if } \|\boldsymbol{X}_{\text{ag}} - \boldsymbol{x}\| > \sqrt{2} \\ 0 & \text{otherwise} \end{cases}$ ▷ Update belief including not found information
26: **end function**

27: **function** $\textsc{move\_source}(\boldsymbol{X}_s(t), \boldsymbol{U}_s(t), P)$
28: $\quad \boldsymbol{X}_s(t+1) = \boldsymbol{X}_s(t) + \boldsymbol{U}_s(t))$
29: $\quad \boldsymbol{U}_s(t+1) \sim P(\boldsymbol{u}'|\boldsymbol{U}_s(t))$
30: **end function**

31: **function** $\textsc{transport\_belief}(b, P)$
32: $\quad b'(\boldsymbol{x}', \boldsymbol{u}') = \sum_{\boldsymbol{u}} P(\boldsymbol{u}'|\boldsymbol{u})b(\boldsymbol{x}' - \boldsymbol{u}, \boldsymbol{u})$
33: **end function**

34: **function** $\textsc{update\_belief}(b, \boldsymbol{X}_{\text{ag}}, h)$
35: $\quad p_h = \sum_{\boldsymbol{x}, \boldsymbol{u}} \mathcal{L}(h|\boldsymbol{X}_{\text{ag}} - \boldsymbol{x})b(\boldsymbol{x}, \boldsymbol{u})$ ▷ Expected probability of the signal
36: $\quad b(\boldsymbol{x}, \boldsymbol{u}) \leftarrow \frac{1}{p_h}\mathcal{L}(h|\boldsymbol{X}_{\text{ag}} - \boldsymbol{x})b(\boldsymbol{x}, \boldsymbol{u})$ ▷ Update belief with Bayes Rule
37: **end function**

38: **function** $\textsc{compute\_heuristic}(b, b', \boldsymbol{X}_{\text{ag}}, \boldsymbol{a})$
39: $\quad Q_{\text{greedy}}(\boldsymbol{a}) = \sum_{\boldsymbol{x}, \boldsymbol{u}} b(\boldsymbol{x}, \boldsymbol{u})Q_{\text{MDP}}^*(\boldsymbol{X}_{\text{ag}} - \boldsymbol{x}, \boldsymbol{u}; \boldsymbol{a})$ ▷ Expected MDP value
40: $\quad (\tilde{b}(\cdot|\boldsymbol{a}), p_f(\boldsymbol{a})) = \textsc{renormalize\_belief}(b', \boldsymbol{X}_{\text{ag}} + \boldsymbol{a})$
41: $\quad$ **for** each observation $h$ **do**
42: $\quad\quad (\tilde{b}(\cdot|\boldsymbol{a}, h), p_h(\boldsymbol{a})) = \textsc{update\_belief}(\tilde{b}, \boldsymbol{X}_{\text{ag}} + \boldsymbol{a}, h)$
43: $\quad\quad \mathcal{H}[b|\boldsymbol{a}, h] = -\sum_{\boldsymbol{x}, \boldsymbol{u}} \bar{b}(\boldsymbol{x}, \boldsymbol{u}|\boldsymbol{a}, h) \log_2 \bar{b}(\boldsymbol{x}, \boldsymbol{u}|\boldsymbol{a}, h)$ ▷ Expected entropy
44: $\quad$ **end for**
45: $\quad \mathcal{H}[b|\boldsymbol{a}] = (1 - p_f(\boldsymbol{a})) \sum_h p_h(\boldsymbol{a}) \mathcal{H}(b|\boldsymbol{a}, h)$
46: $\quad F(\boldsymbol{a}) = wQ_{\text{greedy}}(\boldsymbol{a}) - (1 - w)\mathcal{H}[b|\boldsymbol{a}]$ ▷ Expected value of action $\boldsymbol{a}$ and given weight $w$
47: **end function**

## II. SUPPLEMENTARY MOVIES

### Movies S1-S2. Discrete run-and-tumble search ($\tau_p = 2$ and 25).

Sample search episodes for discrete run-and-tumble setup at persistence times $\tau_p = 2$ and 25, with the agent using hybrid policy with optimal weight. The domain size is $L = 51$ with periodic boundary conditions and lattice spacing $\Delta x = 1$. The target (blue) and the agent (red triangle) both move by one lattice step per time step. The emission rate is $R = 1$ and the likelihood decay length scale is $\lambda = 3$. Olfactory detections are shown as black crosses, and the episode begins with a detection event. The episode ends when the distance between the agent and the target is below $\sqrt{2}$. The grayscale colormap represents the agent's belief over the target state. The inset panel shows the belief over the target velocity together with the true target velocity (blue arrow).

### Movies S3-S4. Comparison between best heuristic and infotaxis at $\tau_p = 2$ and 25.

The setup is the same as for movies S1-S2. The search episode resulting from the best heuristic (corresponding to the best weight $w$ blending infotactic and greedy policies) is shown in the left panel; the search episode resulting from infotaxis (corresponding to $w = 0$) is shown in the right panel.

### Movie S5. Continuous run-and-tumble search ($\mathcal{T}_p = 1$).

Sample search episode for the continuous run-and-tumble setup at persistence time $\mathcal{T}_p = 1$. The domain size is $L = 51$ with periodic boundary conditions and lattice spacing $\Delta x = 1$. The target (blue) moves continuously in the plane, whereas the agent (red triangle) moves by one lattice step per time step. The emission rate is $R = 120$ and the likelihood length scale is $\lambda = 3$. The diffusing odor tracers emitted from the source are shown in light blue (color-coded by the time elapsed from emission). Olfactory detections are shown as black crosses, and the episode begins with a detection event. The episode ends when the distance between the agent and the target is below $\sqrt{2}$. The grayscale colormap represents the agent's belief over the target position. The inset panel shows the belief over the target velocity together with the true target velocity (blue arrow).



\end{document}